\documentclass[preprint,authoryear]{elsarticle}
\usepackage{amssymb}
\usepackage{color,soul}
\journal{Expert systems with applications}
\begin{document}
\begin{frontmatter}
\title{Extracting domain-specific terms using contextual word embeddings}
\author[inst1]{Andraz Repar}
\author[inst1,inst2]{Nada Lavrac}
\author[inst1]{Senja Pollak}
\affiliation[inst1]{organization={Jozef Stefan Institute},
                    addressline={Jamova 39}, 
                    city={1000 Ljubljana},
                    country={Slovenia}}
\affiliation[inst2]{organization={University of Nova Gorica},
                    addressline={Glavni trg 8}, 
                    city={5271 Vipava},
                    country={Slovenia}}
\begin{abstract}
Automated terminology extraction refers to the task of extracting meaningful terms from domain-specific texts. This paper proposes a novel machine learning approach to terminology extraction, which combines features from traditional term extraction systems with novel contextual features derived from contextual word embeddings. Instead of using a predefined list of part-of-speech patterns, we first analyse a new term-annotated corpus RSDO5 for the Slovenian language and devise a set of rules for term candidate selection and then generate statistical, linguistic and context-based features. We use a support-vector machine algorithm to train a classification model, evaluate it on the four domains (biomechanics, linguistics, chemistry, veterinary) of the RSDO5 corpus and compare the results with state-of-art term extraction approaches for the Slovenian language. Our approach provides significant improvements in terms of F1 score over the previous state-of-the-art, which proves that contextual word embeddings are valuable for improving term extraction.
\end{abstract}
\end{frontmatter}


\section{Introduction}
\label{sec:intro}


Automated terminology extraction (ATE) refers to the task of extracting meaningful terms from domain-specific texts. Terms are single-word (SWU) or multi-word units (MWU) of knowledge, which are relevant for a particular domain. Since manual identification of terms is costly and time consuming, ATE approaches can reduce the effort needed to generate relevant domain-specific terms.

Recognizing and extracting domain-specific terms, which is useful in various fields, such as translation, dictionary creation, ontology generation and others, remains a difficult task. ATE has been the subject of research for a long time, but state-of-the-art performance often lags behind other NLP tasks. The most important reason for this is the fact that it is very difficult to provide a clear definition of the nature of domain-specific terms and, as a result,  
annotated datasets for ATE are relatively scarce.
Nevertheless, 
in the recent years, ATE-specific ACTER datasets for English, French and Dutch were released and used in several papers describing state-of-the-art ATE approaches \citep{terryn2020no}, while the recently released RSDO5 dataset for Slovenian \citep{11356/1400} is used in the present paper.

In this paper, we describe a terminology extraction methodology that combines two traditional aspects of ATE with a novel contextual-embedding approach in a machine learning setting. Focusing on the Slovenian language, which is an under-resourced Slavic language with a rich morphology, we conducted the first experiments on a new RSDO5 corpus of term-annotated texts created as part of the RSDO national project. The proposed approach starts with the corpus --- we first analyze the annotated terms in the corpus and study their part-of-speech tags. While traditional systems use a set of pre-defined part-of-speech patterns to identify the initial candidate terms (CTs), we take a different approach and instead define a shallow filter for CTs, which considers only very basic part-of-speech based information. In our approach, we generate three types of features (linguistic, statistical and contextual) and use them to train a linear support vector machine (SVM) classifier. Since the RSDO5 corpus contains four different domains, we train the algorithm on three domains and test its performance on the fourth domain using the standard measures of precision, recall and the F1 score (as in the related work by \cite{rigouts2020termeval}).

The rest of this paper is organized as follows: Section~\ref{sec:related} presents the related work, Section~\ref{sec:datasets} introduces the term-annotated corpus used in the experiments, Section 4 contains a description of the term extraction system, Section 5 contains the results of the experiments, and Section 6 presents the conclusions and ideas for future work.

\section{Related work}
\label{sec:related}

ATE systems were traditionally classified as either statistical, linguistic or a combination of these two approaches. The linguistic approach utilizes the distinctive linguistic aspects of terms, most often their syntactic (i.e. part-of-speech) patterns. On the other hand, the statistical approach takes advantage of term frequencies in a corpus. However, most traditional systems are hybrid, using a combination of the two approaches. For example, \cite{justeson1995technical} first define part-of-speech (POS) patterns of terms and then use simple frequencies to filter the term candidates.  

Many terminology extraction algorithms are based on the concepts of term\-hood and unit\-hood defined by \cite{kageura1996methods}: termhood is “the degree to which a stable lexical unit is related to some domain-specific concepts” and unithood is “the degree of strength or stability of syntagmatic combinations and collocations”. Termhood-based statistical measures \citep{vintar2010bilingual} function on a presumption that a term’s relative frequency will be higher in domain-specific corpora than in the general language, while common statistical measures, such as mutual information \citep{daille1994towards}, are used to measure unithood. These two approaches have been used as a basis of several hybrid systems, such as Termolator \citep{meyers2018termolator} and TermEnsembler \citep{repar2019termensembler}.

However, in the last decade, this division has become too simplistic due to the emergence of new machine learning and deep learning approaches that could not be classified as either linguistic, statistical or hybrid in the traditional sense. The advances in embeddings and deep neural networks have also influenced the terminology extraction field. \cite{amjadian2016local} were one of the first to leverage embeddings for terminology extraction by trying to represent unigram terms as a combination of local and global vectors. Other works involving non-contextual word embeddings include the approaches that either devise a co-training system involving two neural networks to determine whether a term is domain relevant or not \citep{wang2016featureless}, use word embeddings to estimate term similarity in a graph-based ranking system \citep{khan2016term},  employ word embeddings to measure semantic relatedness of term candidates in order to re-rank term candidates generated with traditional term extraction methods \citep{zhang2017semre},  identify term candidates using sequence labeling and word-level and character-level embeddings \citep{kucza2018term}, or devise a nested term extraction classifier with features from various (non-contextual and contextual) word embedding models \citep{gao2019feature}. Contextual word representations, such as eLMO \citep{peters-etal-2018-deep} and BERT \citep{devlin2018bert}, can encode additional information about terms as illustrated by the fact that the winning approach in the TermEval2020 competition \citep{rigouts2020termeval} uses the BERT model. TALN-LS2N \citep{hazem2020termeval}, the winning approach for English and French, uses BERT in a binary classification setting, where a combination of n-grams and a sentence are used as an instance and the classifier needs to determine for each n-gram inside the sentence whether it is a term or not. On the other hand, the winning approach for Dutch \citep{rigouts2020termeval} uses pretrained GloVe word embeddings that are fed into a bi-directional LSTM-based neural architecture. Another well-performing system used in this competition that combines word embeddings with statistical approaches compares the performance of terminology extraction built on an improved TextRank, TFIDF, clustering, and termhood features \citep{puais2020termeval}. Recently, state-of-the-art results for English, French and Dutch have been reported by \cite{lang2021transforming} using a sequence labeling approach.

Another distinct approach is to utilize machine learning with feature engineering. It involves first extracting a set of term candidates, followed by the calculation of various features and training of a machine learning model, where term extraction is treated as a bilingual classification task. Various types of machine learning algorithms have been used, such as decision trees \citep{karan2012evaluation}, rule induction \citep{foo2010using}, k-nearest neighbours \citep{qasemizadeh2014evaluation}, support vector machines \citep{ljubevsic2019kas} and random forest \citep{rigouts2021hamlet}. The last two approaches are particularly relevant for our work. 

The first approach, developed by \cite{ljubevsic2019kas} is the state-of-the-art system for Slovene. It first extracts CTs with the CollTerm tool \citep{pinnis2012term}, which uses a complex language-specific set of term patterns originally developed for the SketchEngine terminology extraction module \citep{fivser2016terminology}. A total of 31 patterns were defined from unigrams up to four-grams and CTs (i.e. lemmatized versions of terms) with a frequency of 3 or more were considered. The resulting term lists were annotated by four annotators as either in-domain, out-of-domain, academic or irrelevant terms (which is a similar setup as used in the ACTER and RSDO datasets). The annotations were then used as training data for a machine learning approach with the following features: term frequency, 5 statistical measures from the SketchEngine terminology module (chi-square, dice, pointwise mutual information, t-score, tf-idf), C-value \citep{frantzi2000automatic}, oversampling (based on instances where the annotators were in agreement), candidate length, average token length, term pattern and context\footnote{Calculated by using a context-based SVM classifier with a linear kernel with features of the classifier being frequencies of tokens occurring in a 3-token window around all the occurrences of a term candidate in the respective document.}. Since some of the statistical measures can be calculated only for multi-word units, they trained separate classifiers for single-word (SWU) and multi-word (MWU) units. They evaluated the models in a cross-validation setting on the Slovene KAS dataset \citep{11356/1198}, obtaining F1 scores of around 0.5 (i.e. when only the \textit{irrelevant} annotation is considered negative and the remaining annotations are treated as valid terms, which is similar to the setting used in our experiments). 

The second approach, which also uses the machine learning paradigm with feature engineering (as we do in our work) was developed by \cite{rigouts2021hamlet}. It first identifies the linguistic patterns of the annotated terms in the ACTER corpus and then uses these patterns to identify term candidates. They generate 177 features in 6 subgroups: shape (e.g., number of tokens in a CT), linguistic (e.g., POS tag of the first token of the CT), frequency (e.g., relative frequency of the CT in a specialized corpus), statistical (various termhood/unithood measures), contextual (e.g., whether the CT occurs between parentheses or right before/after parentheses), variational (number of different variations of the CT). They experimented with several different classification algorithms in the \textit{sklearn} Python library and obtained the best F1 scores with the random forest classifier. In the setup with a held-out test set (3 domains are used for training, one for testing and the experiments are run 4 times, each time with a different test domain) they achieved F1 scores between 0.338 and 0.436 for English, between 0.288 and 0.520 for French and between 0.361 and 0.616 for Dutch.

\section{Corpus}
\label{sec:datasets}
The RSDO5 corpus \citep{11356/1400} was compiled for developing and evaluating ATE methods in Slovene. It consists of 12 texts with 250,000 words and 38,000 manually annotated terms\footnote{Note that this number refers to all occurrences of all terms.}. The corpus texts, published between 2000 and 2019, belong to the fields of biomechanics, linguistics, chemistry, or veterinary science. For each domain, they include: a PhD thesis, a graduate level text book, and a journal article. The entire texts were annotated for terminology and allow for evaluation of methods in terms of precision and recall. Apart from the manually annotated terms, the corpus was automatically annotated, i.e. performing tokenization, sentence segmentation, lemmatization, assigning morphological features and dependency syntax using the Classla pipeline \citep{ljubesic-dobrovoljc-2019-neural}. It is available in the Clarin.si repository.\footnote{https://www.clarin.si/repository/xmlui/handle/11356/1400}

\begin{table}[]
    \centering
    \begin{tabular}{cccccc}
    \hline
         domain & lemma terms & freq=1 & freq=2 & freq$>$2 \\
         \hline
         biomechanics & 1,596 & 891 & 266 & 439 \\
         linguistics & 3,102 & 2,115 & 415 & 532 \\
         veterinary & 1,580 & 880 & 245 & 455 \\
         chemistry & 3,379 & 2,098 & 483 & 798 \\
         \hline
    \end{tabular}
    \caption{Number of lemmatized terms, terms with frequency of 1, terms with frequency of 2 and terms with frequency of more than 2, per domain in the RSDO5 corpus.}
    \label{tab:corpus_stats}
\end{table}

Table \ref{tab:corpus_stats} contains the basis statistics of the terms in the corpus relevant for our term extraction approach. It contains a total of 9,657 unique lemma term forms and a large number of these occur only once or twice in an individual domain of the corpus. 
We have analyzed the annotated terms in the corpus with respect to token and character lengths, POS tags of unigram terms, POS tags of the first token in the terms, POS tags of the last token in the terms, and frequency of different POS tags in the terms.

\begin{figure}[t!]
\includegraphics[width=\textwidth]{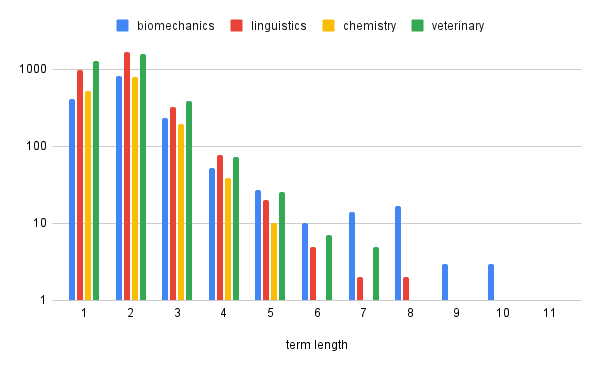}
\caption{Number of tokens in gold standard (lemmatized) terms.}
\label{fig:length}
\end{figure}

\begin{figure}[htbp]
\includegraphics[width=\textwidth]{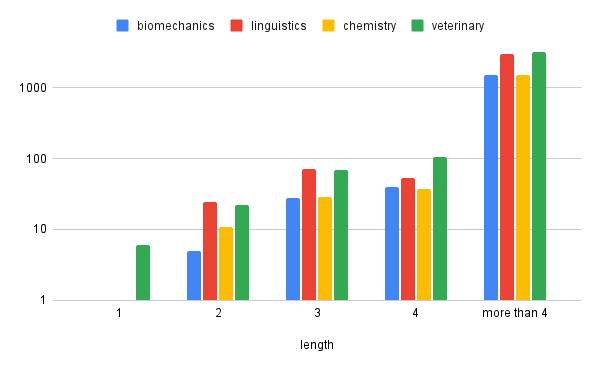}
\caption{Character length of gold standard (lemmatized) terms (including spaces for MWUs).}
\label{fig:character_lengths}
\end{figure}

\begin{figure}[htbp]
\includegraphics[width=\textwidth]{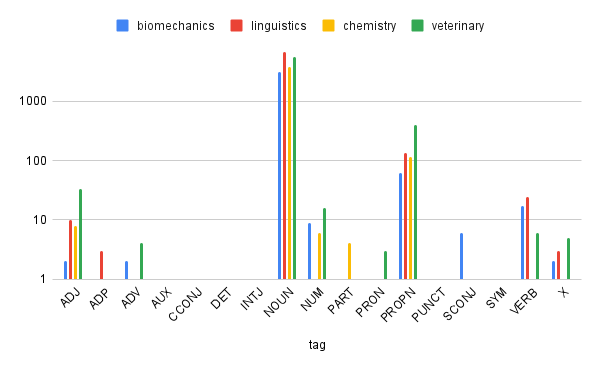}
\caption{POS tag of gold standard unigram (non-lemmatized) terms.}
\label{fig:unigram_pos}
\end{figure}

\begin{figure}[htbp]
\includegraphics[width=\textwidth]{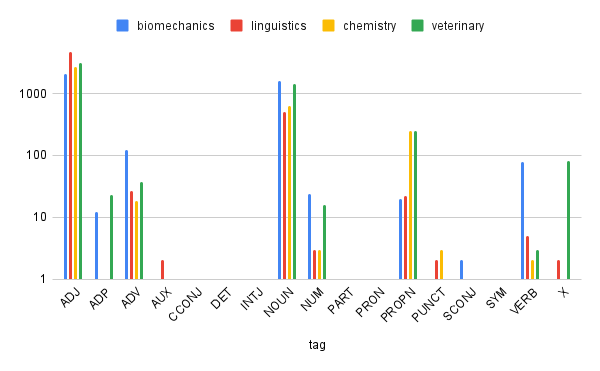}
\caption{First POS tag of annotated (non-lemmatized) terms.}
\label{fig:first_pos}
\end{figure}

\begin{figure}[htbp]
\includegraphics[width=\textwidth]{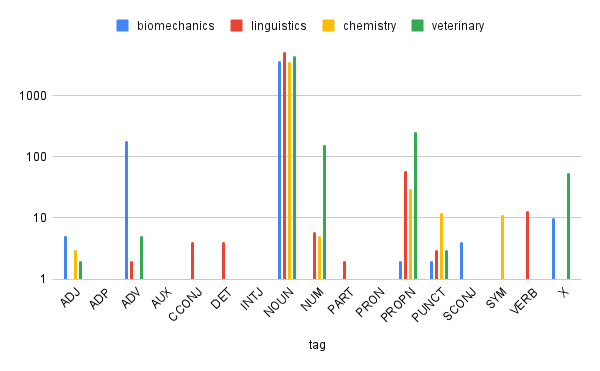}
\caption{Last POS tag of annotated (non-lemmatized) terms (longer than 1 token).}
\label{fig:last_pos}
\end{figure}

\begin{figure}[htbp]
\includegraphics[width=\textwidth]{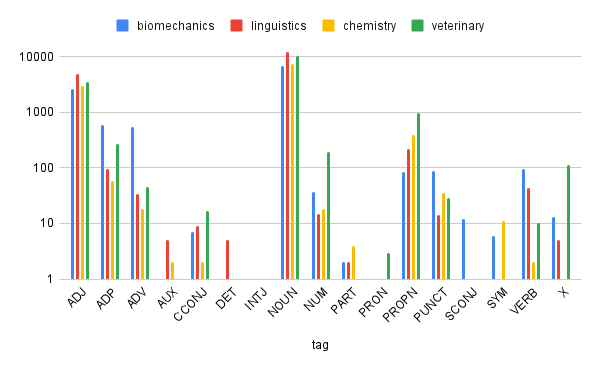}
\caption{Frequency of POS tags that appear in the annotated (non-lemmatized) terms.}
\label{fig:upos_tags}
\end{figure}

As can be observed in Figure \ref{fig:length}, most annotated terms have 4 or less tokens. The longest term per domain has 6 tokens in the chemistry domain, 11 tokens in the biomechanics domain, 10 tokens in the veterinary domain and 8 tokens in the linguistics domain. The vast majority of terms also have more than 4 characters (see Figure \ref{fig:character_lengths}). We can also observe that\footnote{While Figures \ref{fig:length} and \ref{fig:character_lengths} are produced based on the lemmatized term lists, in the counts of Figures \ref{fig:unigram_pos}, \ref{fig:first_pos}, \ref{fig:last_pos} all different appearances of terms are considered  \ref{fig:upos_tags}, due the fact that the syntactic parsing algorithm (Classla) could produce different annotations in different contexts.}:
\begin{itemize}
    \item almost all unigram terms are either nouns (NOUN) or proper nouns (PROPN) (for details, see Figure \ref{fig:unigram_pos}), 
    \item most terms start with either an adjective (ADJ), noun (NOUN) or proper noun (PROPN) (for details, see Figure \ref{fig:first_pos}),
    \item most multi-word unit terms end with a noun (NOUN) or a proper noun (PROPN) (for details, see Figure \ref{fig:last_pos}),
    \item nouns (NOUN) and adjectives (ADJ) are by far the most frequent POS tags that appear in the terms, but adverbs (ADV), adpositions (ADP) and proper nouns (PROPN) can also be found; other POS tags occasionally appear in some terms, but the number of occurrences is low and may in some cases be attributed to errors during the syntactic parsing process (for details, see Figure \ref{fig:upos_tags}).
\end{itemize}

\section{System description}
\label{sec:sys_desc}

\begin{figure}[htbp]
\includegraphics[width=\textwidth]{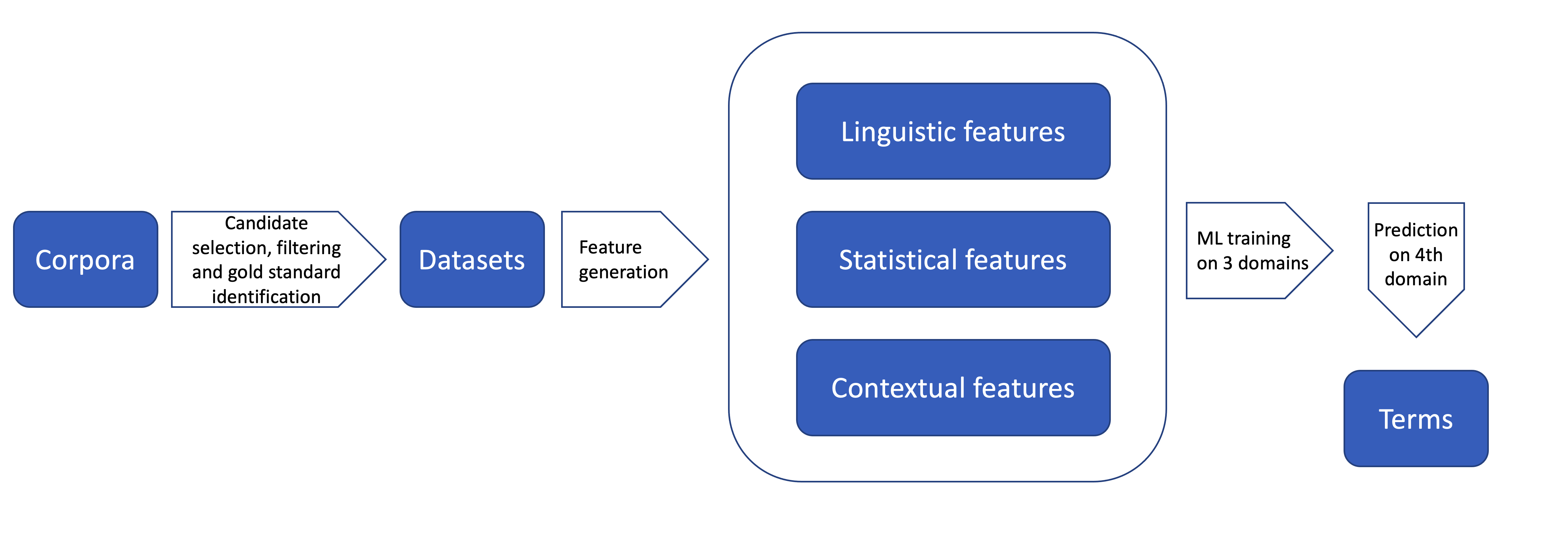}
\caption{Overview of the system. Starting with 4 domain corpora, we generate the datasets by identifying gold standard terms and generating candidates and then calculating 3 types of features. 3 datasets are used to train a machine learning model, while the 4th is used to predict the terms for evaluation.}
\label{fig:flowchart}
\end{figure}

This section describes the architecture of the system. We use a machine learning approach to terminology extraction, which means that for each term candidate, we generate a set of features and then use the corresponding labels for model training. Our approach is similar to the ones developed by \cite{ljubevsic2019kas} and \cite{rigouts2021hamlet} in some respects, but it differs from them in two major aspects: 1) instead of using a pre-defined set of linguistic patterns to identify CTs, we apply 6 filters to all possible n-grams in the input corpus up to a user-defined length \textit{n}, and 2) in addition to linguistic and statistical features, we also employ a set of novel contextual features based on ELMo \citep{peters-etal-2018-deep} contextual word-embeddings. A general overview of the system is depicted in Figure \ref{fig:flowchart}.

\subsection{Dataset generation}
\label{sec:dataset_generation}

In traditional ATE systems, candidate terms (CTs) are usually selected based on a pre-defined list of POS patterns. For example, all adjective-noun (ADJ+NOUN) sequences, such as ``supervised learning'' or ``basic rule'', are considered CTs. Since most terms, in particular high frequent ones, correspond to one of the standard patterns, this allows us to quickly filter out a large number of low-quality CTs. However, defining a POS pattern list that would effectively cover terms across various types of domains is difficult. While some common patterns (e.g., NOUN+NOUN or ADJ+NOUN) can be considered universal, other patterns may be domain-specific and maintaining different pattern lists for many different domains can be cumbersome. Using a traditional pattern-based approach, we also discard potentially valid terms that do not correspond to one of the POS patterns either because they follow a non-standard POS pattern or because they are unusually long. Since most patterns are rarely longer than 4 or 5 tokens (see Figure \ref{fig:length}), longer terms would automatically be discarded.

Instead of relying on a list of POS patterns to identify CTs we apply a shallow filter to all n-grams up to a pre-defined maximum length value. In our case, we set the maximum length to 11, since this is the longest annotated term in the RSDO5 corpus (see Figure \ref{fig:length}). The shallow filter is based on the analysis of the terms in the RSDO5 corpus and describes the general linguistic characteristics of the terms. The rules are described in detail below:

\begin{enumerate}
    \item \emph{Terms have to be longer than 3 characters.} \\
    As evident from Figure \ref{fig:character_lengths}, the vast majority of terms (97.87\% in the biomechanics domain, 96.91\% in the linguistics domain, 97.47\% in the chemistry domain and 97.10\% in the veterinary domain) are longer than 3 characters.
    \item \emph{Only nouns can be single word terms.} \\
    As evident from Figure \ref{fig:unigram_pos}, the vast majority of unigram terms (98.78\% in the biomechanics domain, 99.40\% in the linguistics domain, 99.50\% in the chemistry domain and 98.82\% in the veterinary domain) are either nouns (NOUN) or proper nouns (PROPN)\footnote{Note that we do not distinguish between nouns and proper nouns, because we found that the syntactic parsing process is unreliable when it comes to nouns that can be both regular nouns and proper nouns (such as the word ``commission'' which can be used in the general sense or as part of the proper name ``European Commission'').}. 
    \item \emph{Patterns longer than 1 have to end with a noun (NOUN) or proper noun (PROPN) to be terms.} \\
    As evident from Figure \ref{fig:last_pos}, the vast majority of last POS tags of annotated terms, longer than one token (94.92\% in the biomechanics domain, 99.34\% in the linguistics domain, 99.22\% in the chemistry domain and 95.54\% in the veterinary domain) are either nouns (NOUN) or proper nouns (PROPN).
    \item \emph{Patterns not starting with adjectives (ADJ), adverbs (ADV) or nouns (NOUN, PROPN) are not terms.} \\
    As evident from Figure \ref{fig:first_pos}, the vast majority of first POS tags of annotated terms (96.98\% in the biomechanics domain, 99.74\% in the linguistics domain, 99.72\% in the chemistry domain and 97.51\% in the veterinary domain) are either adjectives (ADJ), adverbs (ADV), nouns (NOUN) or proper nouns (PROPN). 
    \item \emph{If a pattern contains a verb (VERB), a symbol (SYM), a subordinating conjunction (SCONJ), punctuation (PUNCT), a pronoun (PRON), a particle (PART), an interjection (INTJ), a determiner (DET), a coordinating conjuction (CCONJ), an auxiliary verb (AUX) or other (X), it is not a term.} \\
    As evident from Figure \ref{fig:first_pos}, only a small fraction of annotated terms (2.01\% in the biomechanics domain, 0.48\% in the linguistics domain, 0.51\% in the chemistry domain and 1.12\% in the veterinary domain) contain any of these POS tags.
    Despite the fact that adpositions (ADP) and adverbs (ADV) feature in a quite significant number of CTs, in particular in the biomechanics domain (10.16\%)\footnote{They are less prevalent in other domains: 0.74\% in linguistics, 0.69\% in chemistry and 2.05\% in veterinary.}, we discovered during error analysis that a large number of wrongly predicted terms contain adverbs and/or adpositions.
    \item \emph{If term contains a comma or an underscore, it is not a term.} \\

\end{enumerate}

Using these 6 filters, we are able to significantly reduce the number of CTs while maintaining adequate gold standard coverage. The maximum recall per domain is 0.91 for the biomechanics and linguistics domain, 0.86 for the veterinary domain and 0.93 for the chemistry domain, while the number of CTs (including valid gold standard terms) is 12,847 for the biomechanics domain, 22,610 for the linguistics domain, 17,996 for the veterinary domain and 15,417 for the chemistry domain. For details, see Table \ref{tab:dataset_filter} showing the total number of GS terms being filtered out, the upper b.


\begin{table}[]
    \centering
    \begin{tabular}{ccccc}
    \hline
         & GS terms & Filtered out & Max. recall & Candidates  \\
         \hline
         biomechanics & 1,596 & 138 & 0.91 & 12,847 \\
         linguistics & 3,102 & 277 & 0.91 & 22,610 \\
         chemistry & 1,580 & 115 & 0.93 & 15,417 \\
         veterinary & 3,379 & 481 & 0.86 & 17,996 \\
         \hline
    \end{tabular}
    \caption{Dataset filtering effects, where GS denotes gold standard.}
    \label{tab:dataset_filter}
\end{table}

\subsection{Feature construction}
\label{sec:feature_construction}

Similar to traditional ATE systems, we generate linguistic and statistical features and then we also add a third category of features that are based on contextual word embeddings. 

\subsubsection{Linguistic features}

Contrary to some traditional systems, such as the one developed by  \cite{justeson1995technical}, we do not use pre-defined linguistic patterns, but instead generate features based on a set of loosely defined pattern rules utilizing the Universal Dependency (UD) POS tags \citep{10.1162/coli_a_00402}. 

We generate several vectors of UD tags for each CT depending on the UD part-of-speech value as described below. Each vector has a length of 17 corresponding to the number of all possible UD tags:
\begin{itemize}
\item \textit{StartUD}: a one-hot vector where the UD tag of the first token in the CT has a value of 1, while the rest have a value of 0,
\item \textit{EndUD}: a one-hot vector where the UD tag of the last token in the CT has a value of 1, while the rest have a value of 0; in the case of unigram CTs, the \textit{StartUD} and \textit{EndUD} vectors are the same,
\item \textit{AnywhereUD}: a vector where the UD tags that appear anywhere in the CT other than the first and last position have a value of 1, while the rest have a value of 0; in the case of unigram and bigram CTs, this vector has only zero values,
\item \textit{CountOfUD}: a vector indicating the number of occurrences of each UD tag anywhere in the CT.
\end{itemize}

For an illustration of the vectors generated, see Table \ref{tab:feature_example_ling}. Finally, we also generate an additional numeric feature that counts the number of unique POS tags in the term candidate: 
\begin{itemize}
\item \textit{NoUniquePos} 
\end{itemize}

The vectors are concatenated, resulting in a representation of 69 features: 51 features with binary values, and 18 (17 from CountOfUD and 1 NoUniquePos) with numeric values.

\begin{table}[]
    \tiny
    \centering
    \setlength\tabcolsep{1pt}
    \begin{tabular}{llllllllllllllllll}
    \hline
    Vector & ADJ & ADP & ADV & AUX & CCONJ & DET & INTJ & NOUN & NUM & PART & PRON & PROPN & PUNCT & SCONJ & SYM & VERB & X \\
    \hline
    StartUD & 1 & 0 & 0 & 0 & 0 & 0 & 0 & 0 & 0 & 0 & 0 & 0 & 0 & 0 & 0 & 0 & 0 \\
    EndUD & 0 & 0 & 0 & 0 & 0 & 0 & 0 & 1 & 0 & 0 & 0 & 0 & 0 & 0 & 0 & 0 & 0 \\
    AnywhereUD & 1 & 0 & 0 & 0 & 0 & 0 & 0 & 1 & 0 & 0 & 0 & 0 & 0 & 0 & 0 & 0 & 0 \\
    CountOfUD & 1 & 0 & 0 & 0 & 0 & 0 & 0 & 2 & 0 & 0 & 0 & 0 & 0 & 0 & 0 & 0 & 0 \\
    \hline
    \end{tabular}
    \caption{The four vectors generated during feature construction for the term \textit{supervized machine learning} annotated with the following UD tags: ADJ NOUN NOUN.}
    \label{tab:feature_example_ling}
\end{table}

\subsubsection{Statistical features}

For statistical features, we use the \textit{termhood} measure from \cite{vintar2010bilingual}, which is based on the premise that domain-specific terms are used more frequently (in relative terms) in domain texts than in the general language. But instead of calculating a single termhood value, we generate the three core variables (general corpus frequency, domain corpus frequency and term length) from the termhood formula as separate features\footnote{Note that we use the following stoplist to exclude words from frequency calculations: \textit{brez, do, iz, z, s, za, h, k, proti, kljub, čez, skozi, zoper, po, o, pri, po, z, s, na, ob, v, med, nad, pod, pred, za}}:

\begin{itemize}
    \item \textit{TermGenFreq}: the sum of the relative frequencies in a general corpus of individual tokens constituing a CT,
    \item \textit{TermDomFreq}: the sum of the relative frequencies in a domain-specific corpus of individual tokens constituing a CT,
    \item \textit{TermLength}: which corresponds to the length of the CT (i.e. number of tokens).
\end{itemize}

For calculating general corpus relative frequency, we used a word frequency list from the Gigafida 2.0 Slovene reference corpus \citep{krek2020gigafida}, whereas the domain corpus is the training data from the RSDO5 corpus described in Section \ref{sec:datasets}.
The sums of relative frequencies are calculated using the following formula:
\begin{equation}
\sum_{1}^{n} \log{\frac{f_{n}}{N}}
\end{equation}
\noindent where \textit{n} represents the number of tokens in the CT, \textit{f}\textsubscript{n} is the frequency of each token in the CT and \textit{N} is the size of the corpus in tokens. The same formula is used for the calculation of general and domain-specific corpora relative frequencies. The features are concatenated to the linguistic feature vector.



\subsubsection{Contextual features}

To generate contextual features, which is also the main novelty of our approach, we utilize a premise that is similar to statistical termhood measures. Just as termhood suggests that domain-specific terms are used more frequently in domain-specific corpora than in general corpora, so we hypothesize that domain-specific terms are used in different contexts in domain-specific corpora compared to general corpora.

For the calculation of contextual embeddings, we used the eLMO model for Slovenian created by \cite{11356/1277}, which was trained on the Gigafida 2.0 corpus for 10 epochs. General corpus contextual embeddings were calculated for the top 200k most frequent tokens from the publicly available ccGigafida corpus \citep{11356/1035}. To produce the values, the first LSTM layer was used (based on \cite{reimers2019alternative} who report that the middle layer is the best single layer and comparable to concatenating all three layers) to produce the vector values. Each instance of a word has its own vector, based on the context it appears in. These vectors have been averaged, so that each word has only one corresponding vector representing its average context in the general corpus. We then calculate average word embeddings for every word in the domain corpus. To do this, we first tokenize the corpus into sentences and then generate word embeddings for every sentence using the AllenNLP Python library \citep{Gardner2017AllenNLP} using the same ELMo settings as for the generation of the general domain corpus embeddings. We iterate over every word in the sentence and calculate the average embedding of its lemma in the domain corpus by adding up all vectors and dividing them by the number of occurrences of the lemma in the corpus. While for single word terms the average lemma embedding is the final representation, for every multi-word term, we calculate the average embedding by summing up\footnote{{Note that we use the following stoplist to exclude words from contextual embedding calculation: \textit{brez, do, iz, z, s, za, h, k, proti, kljub, čez, skozi, zoper, po, o, pri, po, z, s, na, ob, v, med, nad, pod, pred, za}}} the average lemma embeddings of all tokens in the term and dividing the sum with the number of tokens in the term. All 1,024 dimensions of the resulting vector are then added to the dataset as features. In a similar manner, we then also generate the average general domain term embedding by summing up average lemma embeddings in the general corpus and dividing the sum with the number of tokens in the term. All 1,024 dimensions of the resulting vector (\textit{elmo} feature vector) are again added to the representation feature vector.


Finally, we generate three additional features based on contextual embeddings:
\begin{itemize}
    \item \textit{elmoSim}, which is the cosine similarity of the domain-specific and general term embeddings calculated as described above, the motivation being that since true terms are used in different contexts in domain-specific and general language texts, the similarity between the two vectors would be smaller for true terms compared to expressions that are not valid terms.
    \item \textit{elmoTermSim}, which is the cosine similarity of the domain-specific term embedding and the embedding of a seed term defined by the user\footnote{In the case of the RSDO dataset, we used the domain names as seed terms --- ``veterina'' for the veterinary domain, ``jezikoslovje'' for the linguistics domain, ``biomehanika'' for the biomechanics domain and ``kemija'' for the ``chemistry'' domain.}, the motivation being that true terms would be used in similar contexts to a term representative of the domain.
    \item \textit{elmoStDev}, which is calculated as follows: for every lemma in the domain-specific corpus, we calculate the standard deviation of all its contextual embeddings and then for each term, we sum up the standard deviations of all lemmas and divide the sum with the number of tokens in the term, the motivation being that true terms in most cases appear in similar contexts within domain-specific texts which would result in smaller standard deviation compared to non-valid terms.
\end{itemize}

\section{Experiments and results}

\subsection{Experimental setup}
For model training, we experimented with five algoritms from the \textit{sklearn} Python library. The best F1 score was achieved by a SVM binary classifier with a linear kernel using the default settings (c=1). For detailed results, see Table \ref{tab:algorithms}. Following the setup proposed by \cite{hazem2020termeval}, we use three domains for training and one for testing, which means that we run the experiments four times with a different domain used for testing each time. Evaluation is performed using the standard measures of precision, recall and F1 scores. Precision is calculated as the number of true positive terms divided by the number of all predicted terms, recall is calculated as the number of true positive terms divided by the number of GS terms and F1 score is calculated as the harmonic mean of precision and recall.

\begin{table}[]
    \centering
    \begin{tabular}{lccccc}
    \hline
         Algorithm & bim & ling & chem & vet & average  \\
         \hline
         decision tree & 0.388 & 0.397 & 0.390 & 0.422 & 0.399 \\
         random forest & 0.294 & 0.298 & 0.362 & 0.388 & 0.336 \\
         multiple layer perceptron & \textbf{0.536} & 0.522 & 0.548 & 0.561 & 0.542\\
         logistic regression & 0.535 & 0.568 & \textbf{0.563} & 0.579 & 0.561\\
         support vector machine & 0.530 & \textbf{0.569} & 0.561 & \textbf{0.594} & \textbf{0.564}\\
         \hline
    \end{tabular}
    \caption{F1 scores of various algoritms across domain. Support vector machine has the highest average F1 score.}
    \label{tab:algorithms}
\end{table}

We compare the results to the state-of-the-art for Slovenian \citep{ljubevsic2019kas} (the code for that approach is freely available\footnote{https://github.com/clarinsi/kas-term}), to the LUIZ approach by \cite{vintar2010bilingual} as implemented in \citep{repar2019termensembler}, where a joint list of single and multi-word terms is produced, and to an approach where we use corpus-based patterns similar to \cite{rigouts2021hamlet} instead of our filtering rules\footnote{We first collected all patterns of the terms annotated in the RSDO corpus and then generated candidates that correspond to these patterns.}. For \citep{ljubevsic2019kas}, we trained the MWU and SWU models for all four combinations of training and testing domains and evaluated their performance against the gold standard terms. Minimum frequency was set to 1, which is the same as in our approach. For the LUIZ approach, which does not classify the terms but produces a ranked list of term candidates, we used a cuttoff which corresponded to the number of terms predicted by our approach for a specific domain (i.e. if our approach predicted \textit{n} terms for a domain, we considered the \textit{top n} terms according to the LUIZ score for this domain).


\subsection{Results}

With our approach, we achieve F1 scores of 0.530 for the biomechanics domain, 0.569 for the linguistics domain, 0.561 for the chemistry domain and 0.594 for the veterinary domain (see Table \ref{tab:results_comparison}). These results are comparable with state-of-the-art results for other languages presented in \citep{rigouts2021hamlet} and \citep{lang2021transforming}. Moreover, there is very little variation between domains. 

We also see that our approach exceeds the state-of-the-art by \citep{ljubevsic2019kas} in both precision and recall, which is not surprising given that their method relies heavily on frequency-based features, which work best with high frequency CTs, as well as the more traditional LUIZ approach. Furthermore, a corpus-pattern-based approach similar to \cite{rigouts2021hamlet} exhibits better precision in three of the four domains, but performs worse in terms of F1 score.

\begin{table}[]
    \centering
    \begin{tabular}{lcccc}
    \hline
         Model & Test & Precision & Recall & F1 score  \\
         \hline
         Our approach & bim & 0.650 & \textbf{0.448} & \textbf{0.530} \\
         Pattern approach & bim & \textbf{0.694} & 0.342 & 0.458 \\
         LUIZ & bim & 0.359 & 0.393 & 0.363 \\
         \cite{ljubevsic2019kas} & bim & 0.538 & 0.248 & 0.339 \\
         \hline
         Our approach & ling & 0.672 & \textbf{0.494} & \textbf{0.569} \\
         Pattern approach & ling & \textbf{0.678} & 0.446 & 0.538 \\
         LUIZ & ling & 0.338 & 0.393 & 0.363 \\
         \cite{ljubevsic2019kas} & ling & 0.522 & 0.254 & 0.341 \\
         \hline
         Our approach & chem & 0.691 & \textbf{0.472} & \textbf{0.561} \\
         Pattern approach & chem & \textbf{0.694} & 0.374 & 0.486 \\
         LUIZ & chem & 0.239 & 0.444 & 0.311 \\
         \cite{ljubevsic2019kas} & chem & 0.478 & 0.314 & 0.378 \\
         \hline
         Our approach & vet & \textbf{0.688} & \textbf{0.523} & \textbf{0.594} \\
         Pattern approach & vet & 0.670 & 0.487 & 0.564 \\
         LUIZ & vet & 0.400 & 0.349 & 0.373 \\
         \cite{ljubevsic2019kas} & vet & 0.669 & 0.193 & 0.299 \\
         \hline

    \end{tabular}
    \caption{Precision, recall and F1 score compared to state-of-the-art results for Slovenian. Only the test domain is listed, it can be assumed that the other three domains were used for training. \textit{Our approach} uses the shallow filter described in \ref{sec:dataset_generation}, whereas \textit{Pattern approach} uses corpus-based patterns similar to \cite{rigouts2021hamlet}.}
    \label{tab:results_comparison}
\end{table}

\subsection{Ablation study}
We wanted to analyze the impact of different feature types described in Section \ref{sec:feature_construction} on the final results. One approach, particularly often used in the evaluation of deep learning algorithms, is ablation study \citep{meyes2019ablation}. Analogous to ablation in biology, ablation in machine learning denotes the removal of individual components and studying the effect on the results. In line with this, we have removed the individual feature types from the dataset one-by-one and analyzed the results available in Table \ref{tab:ablation3}. 

We can observe that removing each feature type does reduce the F1 scores in all domains and removing both statistical and linguistic features results in an even bigger drop in F1 scores. When we removed the statistical features but kept linguistic and contextual features, we observed a drop in F1 score performance by 11.70\% in the biomechanics domain, 7.91\% in the linguistics domain, 13.37\% in the chemistry domain and 5.72\% in the veterinary domain. When we removed the pattern features but kept statistical and contextual features, we observed a drop in F1 score performance by 18.30\% in the biomechanics domain, 13.18\% in the linguistics domain, 11.59\% in the chemistry domain and 9.43\% in the veterinary domain. When we removed both statistical and linguistic features but kept contextual features, we observed a drop in F1 score performance by 34.34\% in the biomechanics domain, 27.59\% in the linguistics domain, 25.49\% in the chemistry domain and 17.68\% in the veterinary domain. Using only statistical and pattern features, either together or independently, produces almost no correct predictions.

All three different sets of features contribute to the final result. The drop in F1 score performance when removing linguistic features is somewhat larger than when removing statistical features (with the exception of the chemistry domain) and when we remove both statistical and linguistic features, the results are even worse. However, the results when using just contextual features are still respectable, in particular in terms of precision, which is above 0.630 for all four domains.

\begin{table}[]
    \centering
    \begin{tabular}{cccccccc}
    \hline
         Test domain & C\&P\&S & C\&P & C\&S  & S\&P & C & S & P   \\
         \hline
         bim & \textbf{0.530} & 0.468 & 0.433& 0.206 & 0.348 & 0.000 & 0.003 \\
         ling & \textbf{0.569} & 0.524 & 0.494& 0.174 & 0.412 & 0.000 & 0.000 \\
         chem & \textbf{0.561} & 0.486 & 0.496& 0.247 & 0.418 & 0.000 & 0.001\\
         vet & \textbf{0.594} & 0.560 & 0.538 & 0.089 & 0.489 & 0.000 & 0.000\\
         \hline
    \end{tabular}
    \caption{Ablation study results showing F1 scores with different combinations of feature types (C --- Contextual, P --- Pattern, S --- Statistical).}
    \label{tab:ablation3}
\end{table}

\subsection{Error analysis}

We performed error analysis of the results obtained with the best performing model (i.e. the model based on all three feature types described in Section \ref{sec:feature_construction}). When looking at the false positive predictions in all four domains, we were immediately reminded of the issue we mentioned in the introduction, namely that there is no clear definition of the nature of domain-specific terms. On first look and with the caveat that we are not experts in any of these domains (with the possible exception of linguistics), it would seem that many of the false positives could be valid terms. For example\footnote{Since Slovenian is not a widely known language, we provide English translations in brackets. In addition, please note that while we use canonical forms in the examples for better readability, the system actually produces lemmatized forms.}, \textit{atletska steza (running track)}, \textit{živčni končič (nerve ending)} and \textit{upogibalka (flexor)} in the biomechanics domain, \textit{jezikoslovni model (linguistic model)}, \textit{kodna tabela (code table)} and \textit{nacionalni korpus (national corpus)} in the linguistic domain, \textit{prekurzor (precursor)}, \textit{spektroskopija, (spectroscopy)} and \textit{chronbachov koeficient (cronbach coefficient)} in the chemistry domain and \textit{žvekalka (masseter muscle)}, \textit{stomatitis (stomatitis)} and \textit{nekrotično vnetje (necrotic inflammation)} in the veterinary domain could to an untrained eye look like valid domain-specific terms. We also believe that some of the false positive predicted terms (and many others) would also be useful at least in a ``semi-automatic'' terminology extraction setting, where CTs are first extracted automatically and then evaluated by a domain expert.

In addition to the terms described above, we noted two additional issues among false positives. The first one is general terms/words being predicted as terms, such as leto (year), mesto (city), sistem (system), stopnja (rate), proces (process), sprememba (change), skupina (group), delo (work), zbirka (collection), primer (example), pogoj (condition) etc. The reason for these wrongly predicted terms could again be related to the unclear nature of domain-specific terms, because the gold standard contains some terms that, on first look, do not look much different than the ones mentioned above, such as sila (force), enota (unit), sejem (fair), komora (chamber) etc. The second identified issue is related to the lemmatization algorithm in the Classla pipeline. For example, false positives contains wrongly lemmatized CTs, such as ``regrgrposs'' (lemma of ``REGR\_r\_OsSU'') and ``doogpodlaht'' (lemma of ``D\_O\_podlahti'') or ``mehanovsprejemnik'' (lemma of ``mehano-sprejemniki'') and ``skorajstrokovnjak'' (lemma of ``skoraj-strokovnjakov'').

\section{Conclusion and future work}

This paper presents a machine learning terminology extraction system, which combines elements of traditional termhood- and unithood-based systems with a novel contextual-word-embedding-based approach that takes advantage of the differences in the domain-specific and general language contexts which terms appear in. In addition, we also introduce a novel method of candidate term selection --- instead of being limited to a pre-defined list of part-of-speech patterns, we employ a shallow filter that offers greater flexibility in candidate term selection, in particular when it comes to unseen data and when implementing term extraction in user-facing applications\footnote{This research was conducted as part of a language technology project in Slovenia, where one the applications being in developed is a terminology portal with support for terminology extraction. We believe that non-linguist users would have difficulty defining a comprehensive pattern set and would rather work with verbal constructions such as “my term should start with POS1 and end with POS2”, which correspond nicely with our filtering rules.}. Code and datasets will be made publicly available after publication.

We evaluated the system on a new corpus of term-annotated texts RSDO5 1.0 for Slovenian, created as part of the work in the Slovenian national language technology project RSDO. We compared the results to the existing state-of-the-art approach for Slovenian and were able to improve F1 scores by a considerable margin in all four domains. The novel contextual features based on the eLMO embeddings appear to work well even for low-frequency terms (a well-known issue of traditional statistical methods, many of which are based on frequency counts). In addition, the results also exhibit little variation between test domains.

In terms of time, the most time-consuming part is the calculation of contextual embeddings on the large reference corpus. But since this can be computed only once and reused for further calculations, the system is fairly quick for a corpus of modest size (i.e. 50 to 100 thousand words). E.g., calculating domain-specific embeddings and applying the model is performed in approximately half an hour on a standard laptop without specialized machine learning hardware. This is an acceptable setting for practical applications, e.g. in translation industry or terminology dictionary construction setting. While the current version works only for Slovenian, it would be relatively easy to adapt it to other languages, provided that a suitable general language corpus is available.


In terms of future work, we plan to integrate additional traditional features from the approaches described by \cite{ljubevsic2019kas} and \cite{rigouts2021hamlet} to see if we can further improve our system's performance. We would also like to experiment with a weighted average of all three eLMO layers instead of just a single layer and explore  other contextual word embeddings (e.g., BERT), which generally perform better than eLMO in other NLP tasks. Finally, we also plan to evaluate our approach on other languages, such as English, French and Dutch, which are all available in the ACTER corpus.


\section{Acknowledgements}
The work was partially supported by the Slovenian Research Agency (ARRS) core research programme Knowledge Technologies (P2-0103), ARRS project TermFrame (J6-9372), as well as the Ministry of Culture of Republic of Slovenia project Development of Slovene in Digital Environment (RSDO). The work was also supported by the project Cross-lingual and cross-domain methods for Terminology Extraction and Alignment, a bilateral project funded by the program PROTEUS under the grant number BI-FR/23-24-PROTEUS006

\bibliographystyle{elsarticle-harv} 
\bibliography{cas-refs}





\end{document}